\documentclass{article}

\usepackage{microtype}
\usepackage{amssymb}
\usepackage{graphicx}
\usepackage{booktabs} 
\usepackage{subcaption}
\usepackage{natbib}
\usepackage{bm}

\usepackage{amsmath}
\usepackage{amssymb}
\usepackage{mathtools}
\usepackage{amsthm}

\DeclareMathAlphabet\mathbfcal{OMS}{cmsy}{b}{n}

\usepackage{hyperref}

\usepackage[accepted]{icml2022}

\icmltitlerunning{Model-Free Opponent Shaping}

\begin{document}

\twocolumn[
\icmltitle{Model-Free Opponent Shaping}




\begin{icmlauthorlist}
\icmlauthor{Chris Lu}{ox}
\icmlauthor{Timon Willi}{ox}
\icmlauthor{Christian Schroeder de Witt}{ox}
\icmlauthor{Jakob Foerster}{ox}
\end{icmlauthorlist}

\icmlaffiliation{ox}{Department of Engineering Sciences, University of Oxford, Oxford, United Kingdom}

\icmlcorrespondingauthor{Chris Lu}{christopher.lu@exeter.ox.ac.uk}
\icmlcorrespondingauthor{Timon Willi}{timon.willi@exeter.ox.ac.uk}

\icmlkeywords{Machine Learning, ICML}

\vskip 0.3in
]



\printAffiliationsAndNotice{} 

\begin{abstract}
In general-sum games, the interaction of self-interested learning agents commonly leads to collectively worst-case outcomes, such as defect-defect in the iterated prisoner's dilemma (IPD). 
To overcome this, some methods, such as Learning with Opponent-Learning Awareness (LOLA), shape their opponents' learning process. However, these methods are \textit{myopic} since only a small number of steps can be anticipated, are \textit{asymmetric} since they treat other agents as naive learners, and require the use of \textit{higher-order derivatives}, which are calculated through white-box access to an opponent's differentiable learning algorithm. 
To address these issues, we propose Model-Free Opponent Shaping (M-FOS). M-FOS learns in a \textit{meta-game} in which each meta-step is an episode of the underlying (``inner'') game. The meta-state consists of the inner policies, and the meta-policy produces a new inner policy to be used in the next episode. M-FOS then uses generic \textit{model-free optimisation methods} to learn meta-policies that accomplish long-horizon opponent shaping. 
Empirically, M-FOS near-optimally exploits naive learners and other, more sophisticated algorithms from the literature. For example, to the best of our knowledge, it is the first method to learn the well-known Zero-Determinant (ZD) extortion strategy in the IPD. In the same settings, M-FOS leads to socially optimal outcomes under \textit{meta-self-play}. 
Finally, we show that M-FOS can be scaled to high-dimensional settings. 
Project code is available at: \url{https://github.com/luchris429/Model-Free-Opponent-Shaping}.





\end{abstract}

\section{Introduction}
While much past work in multi-agent reinforcement learning (MARL) has focused on fully-cooperative learning in domains such as Dec-POMDP's \citep{oliehoek_concise_2016} or zero-sum games like Starcraft and Go \cite{silver2017go, vinyals2019alphastar}, these settings only represent a fraction of potential real-world multi-agent environments. General-sum games, which can be neither fully-cooperative nor fully-competitive, describe many domains such as agent-based modeling, social dilemmas, and systems of interacting self-interested agents like self-driving cars. 

Even simple social dilemmas commonly present unique challenges that are not present in single-agent learning \citep{foerster_learning_2018}. For example, in the \textit{IPD} \citep{Axelrod84, Harper_2017}, learning agents that treat their opponents as static parts of the environment typically converge on unconditional mutual defection, which is the globally worst outcome. To avoid such catastrophic outcomes, \citet{foerster_learning_2018} introduce LOLA, which takes into account the opponents' learning step in order to shape their policy. In the self-play setting, LOLA was one of the first methods to discover the reciprocating tit-for-tat (TFT) strategy in the IPD.

However, LOLA and related algorithms, such as SOS \citep{letcher_stable_2019} and Meta Multi-Agent Policy Gradient~\citep[Meta-MAPG]{kim2021meta-mapg}, assume that the opponent is a naive learning (NL) agent, which is often incorrect, e.g. in self-play. Furthermore, to shape their opponents, these methods use second-order derivatives, which are typically high-variance, making learning unstable~\citep{foerster_learning_2018}. Lastly, they are also \textit{myopic} -- they only shape the opponent's next few learning steps, not their long-term development.

To resolve \textit{all} of these issues, we introduce \textbf{M}odel\textbf{-F}ree \textbf{O}pponent \textbf{S}haping (M-FOS). M-FOS is a general meta-learning algorithm that learns over multiple opponent-learning steps \textit{without requiring a model of its opponent's underlying learning algorithm}. 

The core of M-FOS is a \textit{meta-game} in which each meta-step is an episode of the underlying (``inner'') game. The meta-state consists of the inner policies, and the meta-policy produces a new inner policy to be used in the next episode. M-FOS then uses generic \textit{model-free optimisation methods}, rather than approaches that require higher-order derivatives, to learn meta-policies that accomplish long-horizon opponent shaping.
Furthermore, training M-FOS in \textit{meta-self-play} allows mutual opponent shaping without causing the kind of infinite regress typically caused by ever higher-order learning awareness~\citep{foerster_learning_2018}. 

However, since M-FOS is naively model-free, the meta-self-play setting reduces to independent learning, which is highly initialisation-dependent and unstable in general-sum settings. To mitigate this, we introduce a training schedule inspired by Cognitive Hierarchies (CH) \citep{camerer2003cognitivehierarchies}. With this schedule, M-FOS learns to reciprocate with itself in the meta-game, even achieving higher scores than LOLA in self-play.

For low-dimensional games, M-FOS directly learns policy updates by taking policies as input and outputting the next policy as an action.
However, directly inputting and outputting policies does not scale to higher-dimensional games.
We introduce a variant of M-FOS that takes past trajectories as inputs to meta-learn across its opponent's learning steps. We then demonstrate that, even in social dilemmas with temporally-extended transition dynamics, M-FOS still manages to shape naive learners and find mutually beneficial solutions in meta-self-play.

In the experiment section, we show that M-FOS can exploit naive learners much better than a set of widely used general-sum learning algorithms \citep{foerster_learning_2018, kim2021meta-mapg}. In the IPD, M-FOS discovers a famous strategy known as ZD extortion \citep{press_iterated_2012} when playing against NL agents. Notably, unlike other algorithms, it does so \textit{without} access to the opponent's underlying learning algorithm. 
M-FOS even learns to exploit other general-sum algorithms, such as LOLA. 

\section{Related Work}


\textbf{Opponent Shaping: }
Several methods recognise that their current actions influence the future policies of learning opponents and take advantage of this to ``shape'' an opponent's policy to desirable values. Most of these works assume white-box access to an opponent's learning algorithm and reward in order to take higher-order derivatives through an opponent's update \citep{foerster_learning_2018, letcher_differentiable_2019, kim2021meta-mapg, willi2022cola}. Such updates are also myopic since anticipating many steps is intractable. In self-play, these methods inconsistently assume that their opponent is a naive learner. M-FOS does not assume white-box access to an opponent's underlying learning algorithm or reward, does not require higher-order derivatives (which are often high-variance), can shape opponents across a large number of updates, and is consistent in self-play.

\textbf{Opponent Modeling:}
Much work in MARL has focused on the idea of \textit{opponent modeling} in which an agent attempts to model some aspect of the policy of other agents in the environment. This includes explicitly modeling opponent policies \citep{mealing2017opponent}, modeling opponent intentions \citep{raileanu2018modelingothers}, classifying opponent strategies \citep{weber2009datamining, synnaeve2011bayesian}, and modeling an opponent's nested beliefs \citep{wen2019probabilistic}. LILI \citep{xie2020learninglatent} models an opponent's high-level latent strategy from local observations with a latent dynamics model rather than explicitly modeling the opponent's policy. Other work \citep{chakraborty2014multiagent} has also considered learning effective policies in the presence of opponents that have memory.
However, these methods are not capable of actively \textit{shaping} their opponents' learning dynamics, thus they do not address the issue that we address in this paper. 

\textbf{Multi-Agent Meta-Learning:}
M-FOS is a form of multi-agent meta-learning where the meta-policy is parameterized by a neural network. Existing multi-agent meta-learning methods, such as Meta-Policy Gradient (Meta-PG) \citep{alshedivat2018mpg}, Meta-MAPG \citep{kim2021meta-mapg}, and Learning to Exploit (L2E) \citep{wu2021l2e} instead parameterize the meta-policy using a method similar to that of Model-Agnostic Meta-Learning (MAML) \citep{finn2017maml}, in which they learn initial parameters and meta-learn across their own gradient updates. While this type of meta-learning can adapt to any task at test time in single-agent settings \citep{xiong_2021_meta_consistency}, in multi-agent settings, the calculated gradient may not correspond to a direction of improvement as the updates of other agents change the underlying dynamics. 
Rather than being restricted to a gradient update within the episode, M-FOS allows for arbitrary meta-policies that can carry out long horizon opponent shaping.

\section{Background}


A \textbf{partially observable stochastic game} ~\citep[POSG]{kuhn1953_posg} consists of a tuple $\mathcal{M}_{n}=\langle\mathbfcal{I}, \mathcal { S }, \mathbfcal{A}, \bm{\Omega}, \mathcal{O}, \mathcal{P}, \mathbfcal{R}, \gamma\rangle$, where $\mathbfcal{I}=\{1, \ldots, n\}$ denotes a set of $n$ agents, $\mathcal{S}$ denotes the state space, $\mathbfcal{A}=\times_{i \in \mathcal{I}} \mathcal{A}^{i}$ represents the joint action space, $\bm{\Omega}=\times_{i \in \mathcal{I}} \bm{\Omega}^{i}$ the joint observation space, $\mathcal{P}: \mathcal{S} \times \mathbfcal{A} \mapsto \mathcal{S}$ denotes the transition probability function, $\mathcal{O}: \mathcal{S} \times \mathbfcal{A} \times \bm{\Omega} \rightarrow[0,1]$ is the observation function, $\mathbfcal{R}=\times_{i \in \mathcal{I}} \mathcal{R}^{i}$ represents the set of reward functions of all agents, and $\gamma \in[0,1)$ denotes the discount factor. At each timestep $t$, every agent samples an action from its stochastic policy, $a_{t}^{i} \sim \pi^{i}\left(\cdot \mid o^{i}_{t}, \phi^{i}\right)$, where the joint actions at timestep $t$ are 
$\bm{a}_{\bm{t}}=\left\{a_{t}^{i}, \bm{a}_{\bm{t}}^{-i}\right\}$ and $\bm{-i}$ stands for all agents except $i$. The policy is parameterized by $\phi^{i}$. Given the joint actions and the current state, each agent receives their respective reward $r_{t}^{i}=\mathcal{R}^{i}\left(s_{t}, \bm{a}_{\bm{t}}\right)$. Finally, a new state is sampled $s_{t+1} \sim\mathcal{P}\left(\cdot \mid s_{t}, \bm{a}_{\bm{t}}\right)$.

Popular special cases of POSGs are \textit{fully observable} stochastic games where all agents observe the full state at each time step; single-player, i.e. $\mathbfcal{I} = \{1\}$, partially observable Markov decision processes (POMDPs), and MDPs, where the single player observes the full state at each time step. 








\section{Model-Free Opponent Shaping}

Typically opponent-shaping methods are based on MAML-like approaches~\cite{foerster_learning_2018,letcher_stable_2019,kim2021meta-mapg} and use higher-order derivatives to directly shape the opponents' parameter update, which requires white-box access to their differentiable learning algorithm. Furthermore, opponent shaping typically creates a conceptual problem: To shape an opponent, an algorithm needs to specify the learning behaviour of other agents in the environment, e.g. by treating them as \textit{naive learners}, as is done in LOLA~\citep{foerster_learning_2018}. This leads to a fundamental inconsistency in self-play when two of these agents are training together. Even though they are both \textit{opponent shaping} they treat each other as \textit{naive learners}, which can lead to undesired outcomes~\citep{letcher_differentiable_2019}.
Lastly, most opponent-shaping methods only shape the next learning steps instead of considering longer horizons.

Opponent shaping can be formulated as a meta-game, in which the meta-state consists of the policies of all agents, a meta-step is an inner episode, the reward is the inner return, and the meta-action is choosing the next inner policy, where ``inner'' refers to the underlying game.
The \textit{key insight} underlying Model-Free Opponent Shaping (M-FOS) is that we can resolve all of the issues above by directly training meta-policies using model-free optimisation methods that are appropriate for sequential settings, rather than relying on MAML-like approaches.
\begin{algorithm}[tb]
   \caption{General M-FOS}
   \label{alg:mfos}
\begin{algorithmic}[1]
   \STATE Initialize M-FOS parameters $\theta$.
   \WHILE{true}
   \STATE Initialize agents' parameters $\phi^{i}_{0}, \bm{\phi}^{-i}_{0}$.
   \FOR{$t=0$ {\bfseries to} $T$}
   \STATE Reset environment
   \STATE Gather trajectories $\tau_{\bm{\phi}}$ given $\phi^{i}_{t}, \bm{\phi}^{-i}_{t}$
   \STATE Update $\bm{\phi}^{-i}_{t+1}$ according to respective learning algorithms
   \STATE Update $\phi^{i}_{t+1}$ according to meta-policy $\pi_{\theta}$
   \ENDFOR
   \STATE Update $\theta$
   \ENDWHILE
\end{algorithmic}
\end{algorithm}


We formally construct the meta-game as a POMDP $\langle \mathcal { \bar{S} }, \mathcal{\bar{A}},\Omega, \mathcal{\bar{O}}, \mathcal{\bar{P}}, \mathcal{\bar{R}},  \bar{\gamma}\rangle$ over an underlying POSG $\mathcal{M}_{n}$. The meta-game is partially observable because we do not assume full access to the opponents' parameters. 
The M-FOS meta-agent controls agent $i \in \mathbfcal{I}$ in the underlying POSG $\mathcal{M}_{n}$. The state space $\bar{S}$ of the meta-game consists of the policy parameters of the agents in the underlying POSG, $\bar{s}_t = (\phi^i_{t-1}, \bm{\phi}^{-i}_{t-1}) \in \mathcal{\bar{S}}$. The meta-agent's action space consists of agent $i$'s policy, for example outputting a conditioning vector or setting agent $i$'s policy parameters directly, $\bar{a}_t = \phi^i_{t} \sim \pi_{\theta}(\cdot\mid \bar{o}_t)$. Here the meta-policy is parameterized by $\theta$. The meta-agent receives observation $o_t \in \Omega$ with probability $\mathcal{\bar{O}}(\bar{o}_t\mid\bar{s}_t, \bar{a}_{t})$. After each meta-episode, the scalar reward is $\bar{r}_t = \sum_{k=0}^K r^i_{k}(\phi^i_t, \bm{\phi}^{-i}_t)$, where $K$ is the length of the inner episode (i.e. the \textit{reward} in the meta-game at each step is the inner \textit{return}). Finally, a new meta-state is sampled from a stochastic transition probability function, $\bar{s}_{t+1} \sim \bar{P}(\cdot \mid \bar{s}_t, \bar{a}_t)$. $\bar{P}(\bar{s}_t, \bar{a}_t)$ is stochastic since, in general, the update function for any agent can be stochastic, $\phi^{j}_{t+1} \sim h(\cdot \mid \phi^{j}_{t})$. For example, when agent $j$ updates their parameters with policy gradients. Consequently, the trajectory is denoted as $\bar{\tau}_{\theta}:=\left(\bar{o}_{0}, \bar{a}_{0}, \bar{r}_{0}, \ldots, \bar{r}_{T}\right)$, where T is the length of the meta-episode. We train the meta-policy to maximise the expected return per meta-episode $J = \sum_{t=0}^T \bar{r}^i_t(\phi^i_t, \bm{\phi}^{-i}_t)$. Crucially, rather than relying on higher-order derivatives, M-FOS uses \textit{model-free optimisation} methods to directly train a meta-policy. In the Section~\ref{sec:experiments} we show that PPO \citep{schulman2017ppo, pytorch_minimal_ppo} and Genetic Algorithms \citep{such2017deepneuroevolution} work well in this general meta-learning framework. 


\subsection{M-FOS Self-Play}
By doing model-free optimisation in the meta-game, we no longer require higher-order derivatives and also can learn strategies that engage in long-horizon opponent shaping. Next, we also address the issue of symmetry and consistency by introducing \textit{meta-self-play}. When using meta-gradient approaches for opponent shaping, attempts of consistent self-play lead to \textit{infinite recursions}, since each agent differentiates through the learning step of the other agent and so on. 
In contrast, since M-FOS is entirely model-free, \textit{meta-self-play} between two M-FOS agents simply corresponds to learning in a general-sum game, where model-free methods can be applied without causing infinite regress. 

One challenge is that independent learning in general-sum settings is highly initialisation dependent and unstable, which is undesirable for a principled method. Furthermore, in general-sum games there are often multiple nash equilibria, which implies that a stricter way to select equilibria would be desirable.

One such way to select ideal equilibria is called the tracing procedure \cite{harsanyi1988general}. The tracing procedure is based on the idea of each agent having a (common) initial bayesian prior over how a rational agent would behave in a general-sum game. The agents then repeatedly update their policies against each other until convergence, initialising from this prior. The exact procedure is impractical in high-dimensional function approximation, but we use it as inspiration to create a similar self-play approach.

This is implemented via a parameter $\lambda$ that corresponds to the probability of an M-FOS agent being paired with a naive learner rather than another M-FOS agent.  
By setting $\lambda=1$ at the beginning of training, we ground the training to an approximate best-response to NL, while annealing it to $\lambda=0$ allows us to transition to self-play over the course of training gradually.
We anneal $\lambda$ slowly enough, such that the M-FOS agents are always playing near optimally for the given distribution. 

\section{Experimental Setup}
\subsection{Environments}




\textbf{IPD:} The prisoner's dilemma is one of the most widely-studied and important general-sum games, with applications in evolutionary biology, economics, politics, sociology, and other fields \citep{rapoport1965prisoner}. In the prisoner's dilemma, agents can choose to cooperate (C) or defect (D) against each other, with the payouts of the result being presented in Table \ref{table:pd_payoff}.

\begin{table}[hbt!]
\centering
\caption{Payoff Matrix for the Prisoner's Dilemma}
\begin{tabular}{c|c|c|}
  & C        & D        \\ \hline
C & (-1, -1) & (-3, 0)  \\ \hline
D & (0, -3)  & (-2, -2) \\ \hline
\end{tabular}
\label{table:pd_payoff}

\end{table}

A common extension of the prisoner's dilemma is the IPD, in which the prisoner's dilemma is played repeatedly, with players able to observe their opponent's past decisions. Axelrod \citep{Axelrod84} famously held an IPD tournament where a strategy known as TFT, in which a player copies the other player's last move, was popularized. 

Despite decades of previous study of the IPD, \citet{press_iterated_2012} made a surprising mathematical discovery that dramatically changed our understanding of the game: There exist fixed policies, called ZD extortion strategies, that dominate any learning opponent. More specifically, ZD extortion enforces a linear relationship between the two agents' rewards that \textit{disproportionately} benefits the extortioner (see Figure \ref{fig:LABR}). However, it is still in a learning agent's best interest to cooperate against extortion despite the fact that it benefits the extortioner more. In principle, an agent could overcome the extortion by being ``meta''-aware that the opponent can change their policy and punish an extorting opponent.

\textbf{Iterated Matching Pennies:} Iterated Matching Pennies (IMP) is an iterated matrix game like the IPD but is zero-sum. In IMP, agents can play ``Heads'' or ``Tails'' and get payouts according to Table \ref{table:mp_payoff}. 

\begin{table}[hbt!]
\centering
\caption{
Payoff Matrix for Matching Pennies}

\begin{tabular}{c|c|c|}
  & H        & T        \\ \hline
H & (+1, -1) & (-1, +1)  \\ \hline
T & (-1, +1)  & (+1, -1) \\ \hline
\end{tabular}
\label{table:mp_payoff}
\end{table}


\textbf{Chicken Game:} The Chicken Game is a stochastic matrix game. Agents can either Swerve (C) or head Straight (D). While agents can gain a small reward by heading straight against a swerving opponent, they incur a large negative cost if they both head straight. It is often used in political science and economics to describe brinksmanship scenarios in which there is a threat of mutually assured destruction \cite{rapoport1966chicken}. 

\begin{table}[hbt!]
\centering
\caption{Payoff Matrix for the Chicken Game}
\begin{tabular}{c|c|c|}
  & C        & D        \\ \hline
C & (0, 0) & (-1, +1)  \\ \hline
D & (+1, -1)  & (-100, -100) \\ \hline
\end{tabular}
\label{table:chicken_payoff}
\end{table}

\textbf{Matrix Game Setup:} In this paper, we directly calculate the value function rather than repeatedly sampling actions. We initialize all policies randomly (except for M-MAML) by taking the sigmoid of samples from the standard normal distribution. We then calculate the value functions for both policies and update them for $T=100$ steps.

\textbf{Coin Game:} The Coin Game is a multi-agent grid-world environment that simulates social dilemmas like the IPD but with high dimensional dynamic states. First proposed by \citet{lerer_2017_coingame}, the game consists of two players, labeled red and blue respectively, who are tasked with picking up coins, also labeled red and blue respectively, in a 3x3 grid. If a player picks up any coin by moving into the same position as the coin, they receive a reward of $+1$. However, if they pick up a coin of the other player's color, the other player receives a reward of $-2$. Thus, if both agents play greedily and pick up every coin, the expected reward for both agents is $0$. 

\begin{figure}[hbt!]
    \centering
    \includegraphics[width=0.4\textwidth]{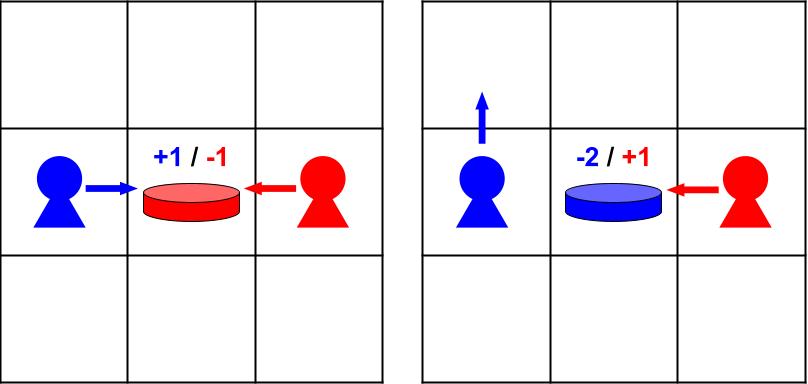}
    \caption{Illustration of Coin Game
    \vspace{-1.0\baselineskip}
    }
    \label{fig:CoinGame}
\end{figure}

\subsection{Baseline Comparisons}

\textbf{Naive Learning (NL):} Naive learners assume that other agents are part of the environment and are static between episodes. Thus, between each episode, naive learners  perform the following update with learning rate $\alpha$:
\vspace{-1.0\baselineskip}

$$\phi^i_{t+1} = \phi^i_t + \alpha \nabla_{\phi^i_t}\mathcal{R}^i(\phi_t^i, \phi_t^{-i})$$

In reinforcement learning, this is often approximated with a sample-based approach. In our experiments, in the Coin Game, the NL uses PPO, \cite{schulman2017ppo} which modifies this by clipping the update.
In matrix games, we can directly perform gradient ascent without sampling because the exact value $\mathcal{R}^i$ is differentiable. 



\textbf{Learning with Opponent Learning Awareness (LOLA):} LOLA assumes that other agents are naive learners and perform the gradient step performed above. LOLA takes a gradient through the opponent's update function to shape the opponent. 
\begin{align}
    \phi^i_{t+1} & = \phi^i_t + \alpha^i \nabla_{\phi^i_t}\mathcal{R}^i(\phi_t^i, \phi_t^{-i} + \Delta \phi^{-i}_t) \\ 
    \Delta \phi^{-i}_t & = \alpha^{-i} \nabla_{\phi^i_t}\mathcal{R}^{-i}(\phi_t^i, \phi_t^{-i}) \nonumber
\end{align}


\textbf{Multiagent Model-Agnostic Meta-Learning:} We introduce a new baseline, Multiagent MAML (M-MAML), which is inspired by Meta-Multiagent Policy Gradient~\citep[Meta-MAPG]{kim2021meta-mapg}. Meta-MAPG and M-MAML operate in a similar setting to M-FOS in that they meta-learn over multiple opponent learning updates. However, instead of learning an update function, they learn initial parameters. They then meta-learn over their own gradient updates (much like MAML \cite{finn2017maml}) as well as the gradient updates of their opponents.
Meta-MAPG and M-MAML optimize the following:
\begin{align}
    & \max_{\phi_0^i} \mathbb{E}_{p(\phi_0^{-i})}[\sum_{t=0}^{t=T}\mathcal{R}^i(\phi_t^i, \phi_t^{-i})], \\
    & \phi_{t+1}^i = \phi_t^i + \alpha^i \nabla_{\phi_t^i} \mathcal{R}^i(\phi_t^i, \phi_t^{-i}) \nonumber \\ 
    & \phi_{t+1}^{-i} = \phi_t^{-i} + \alpha^{-i} \nabla_{\phi_t^{-i}} \mathcal{R}^{-i}(\phi_t^i, \phi_t^{-i}) \nonumber
\end{align}
I.e., the methods only optimize initial policy parameters, assuming that all agents are naive learners.

Meta-MAPG expands the objective into multiple learning terms to perform policy-gradient updates. However, we do not directly compare to Meta-MAPG because it only scales to $T=7$ meta-steps in the IPD, not $T=100$.
Instead, we use the exact value function and exact gradients allowing our baseline (M-MAML) to scale to meta-episodes consisting of $100$ inner episodes. 

\subsection{M-FOS Implementation Details}

Although M-FOS can be applied to any POSG, different settings allow for very different architectures. Below we describe the architectures we use for basic matrix games and the higher-dimensional coin game. 

\textbf{Matrix Games:} In the matrix game environments we allow M-FOS to observe the full state, which is the concatenation of the policies played last timestep $o_t = s_t = (\phi_{t-1}^i, \phi_{t-1}^{-i})$. Because all of our evaluated opponents (including M-FOS itself) only make updates according to the current state, this turns the induced POMDP into an MDP. Because the inner policy can be fully expressed with very few parameters, we can directly output the parameters, turning the MDP into a basic continuous control problem. Because of this, we model the M-FOS meta-agent as a simple feed-forward neural network parameterized by $\theta$ that takes in the state and outputs a distribution over the next policy.
\begin{align}
    & \max_{\theta} \mathbb{E}_{p(\phi_0^{-i}, \phi_0^{i})}[\sum_{t=0}^{t=T}\mathcal{R}^i(\phi_t^i, \phi_t^{-i})], \\  
    & \phi_{t+1}^i \sim \pi_{\theta}(\cdot\mid\phi_{t}^i, \phi_{t}^{-i}) \nonumber \\
    & \phi_{t+1}^{-i} = f(\phi_t^i, \phi_t^{-i}) \nonumber
\end{align}
This can be seen as being related to hypernetwork meta-learners since it directly outputs the weights of another (very simple) model \citep{zhmoginov2022hypertransformer}.

We optimize the meta-policy using both Genetic Algorithms \cite{such2017deepneuroevolution}, and PPO \cite{schulman2017ppo, pytorch_minimal_ppo}, and report the best of both. A detailed breakdown of the performance of each can be found in the Appendix~\ref{sec:detailed-results}.

\textbf{M-FOS in Coin Game:} Here, M-FOS does not directly observe the opponent's policy parameters but only the effects of their past actions. The opponent is parameterized by a convolutional neural network and, as a naive learner, is trained using PPO. 
M-FOS's inner policy is parameterized by a convolutional recurrent neural network that takes in an observation as input along with a conditioning vector from the meta-policy. We require the inner policy to be recurrent to respond to and shape the opponent's policy. The hidden state of the recurrent neural network is reset each episode. 
M-FOS's meta-policy is parameterized by a convolutional recurrent neural network that processes the batch of trajectories from the last episode and outputs a conditioning vector, used in the next episode. 
Using PPO, the inner policy and the meta-policy parameters are trained end-to-end to maximise the expected discounted meta-return.

\begin{figure*}[t!]
    \centering
    \includegraphics[width=0.9\textwidth]{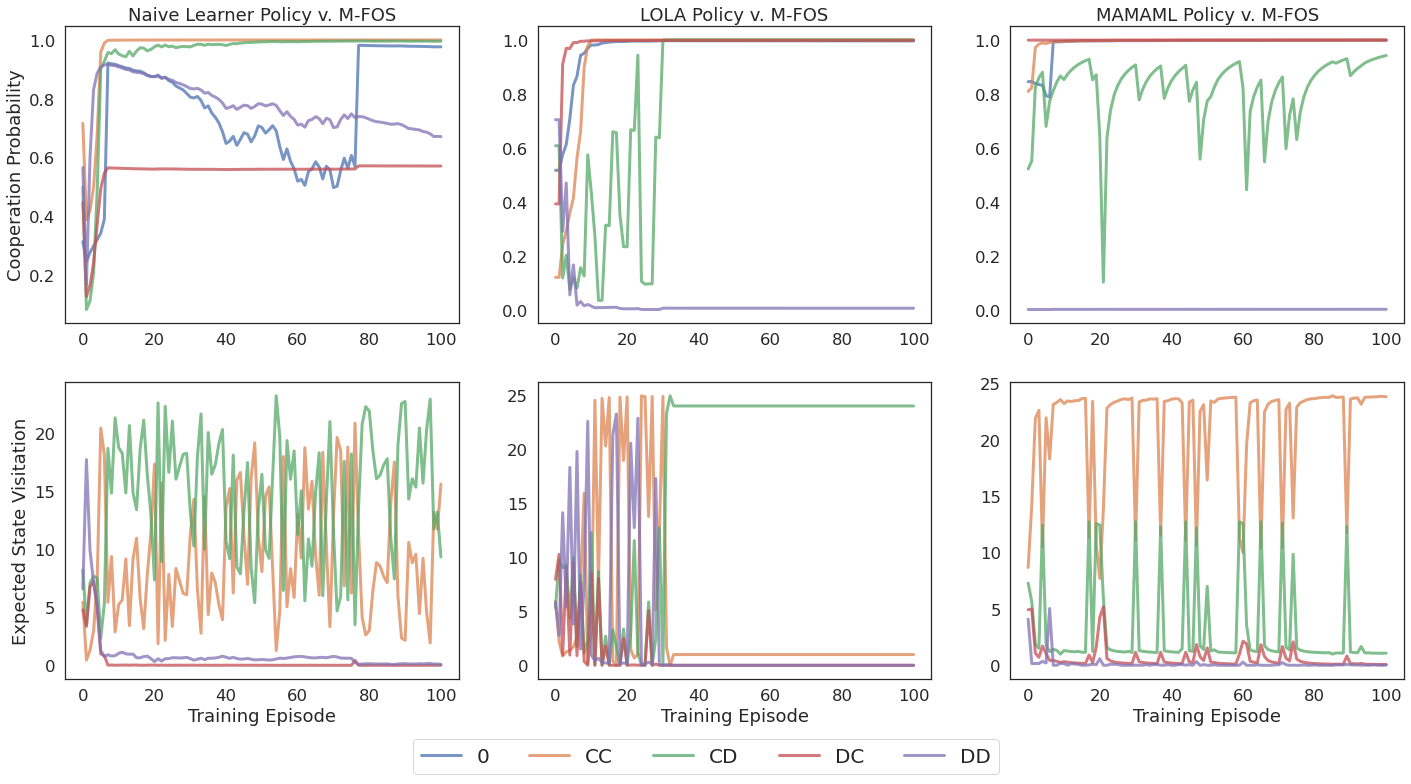}
    \vspace{-10pt}
    \caption{Visualisations of a run of a meta-episode of each learner against M-FOS. Notice how the opponents' policies are shaped into cooperating, resulting in state visitations that are beneficial to the M-FOS agent.}
    \label{fig:IPD_plots}
\end{figure*}

\section{Results}
\label{sec:experiments}

\begin{figure*}[t!]
    \centering
    \begin{subfigure}{.22\textwidth}
      \centering
      \includegraphics[width=.99\linewidth]{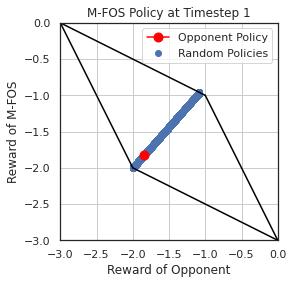}
      \caption{}
      \label{fig:MFOS_T1}
    \end{subfigure}
    \begin{subfigure}{.22\textwidth}
      \centering
      \includegraphics[width=.99\linewidth]{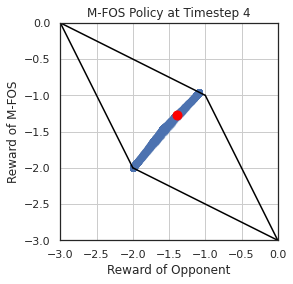}
      \caption{}
      \label{fig:MFOS_T4}
    \end{subfigure}
    \begin{subfigure}{.22\textwidth}
      \centering
      \includegraphics[width=.99\linewidth]{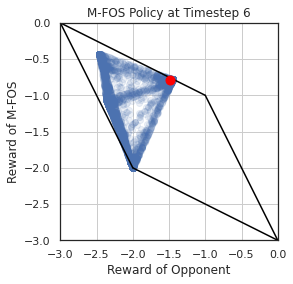}
      \caption{}
      \label{fig:MFOS_T6}
    \end{subfigure}
    \begin{subfigure}{.22\textwidth}
      \centering
      \includegraphics[width=.99\linewidth]{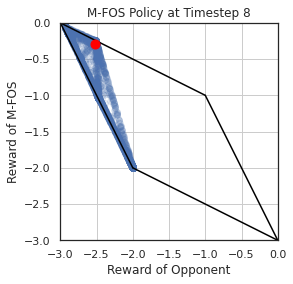}
      \caption{}
      \label{fig:MFOS_T8}
    \end{subfigure}

    \caption{Visualisations of M-FOS shaping a naive learner. The area denoted by the black lines represents the episode's possible rewards. The blue points represent the possible payoffs of a naive learner against the M-FOS policy at that timestep. (a)-(b) M-FOS begins by playing TFT until the opponent is sufficiently cooperative. (c)-(d) M-FOS then repeatedly switches between an extortion-like policy (c) and a defecting policy (d), making the NL oscillate.
    \vspace{-1.5\baselineskip}
    }
    \label{fig:IPD_Viz}
\end{figure*}

\subsection{Iterated Prisoner's Dilemma}

In a round-robin tournament in which algorithms train against each other in a head-to-head matchup, M-FOS vastly outperforms all other learning methods in the IPD. Notably, it is the only algorithm to achieve scores better than mutual cooperation ($-1$), and it does so against all opponents, excluding itself. Similarly, it is the only algorithm for which one of its opponents performs worse than mutual defection ($-2$), and it does so against both naive learners and LOLA.

\begin{table}[hbt!]
\caption{Head-to-head rewards of each learning algorithm in the Iterated Prisoner's Dilemma. }
\label{table:IPD_results}
\begin{tabular}{l|llll}
      & M-FOS          & NL              & LOLA            & M-MAML \\ \hline
M-FOS  & \textbf{-1.01} & \textbf{-0.51} & \textbf{-0.73} & \textbf{-0.67}     \\ \hline
NL     & -2.14          & -1.98           & -1.52           & -1.28     \\ \hline
LOLA   & -2.09          & -1.30           & -1.09           & -1.04     \\ \hline
M-MAML & -1.86             & -1.25              & -1.15              & -1.17     \\ \hline
\end{tabular}
\vspace{-1.0\baselineskip}
\end{table}



\textbf{M-FOS v. Naive Learner:} Against an NL agent, M-FOS gets an average score of $-0.51$, while the NL agent gets an average score of $-2.14$. This is a far more advantageous result than LOLA achieves ($-1.30$/$-1.52$), even though LOLA has a perfect learning model of its opponent and can take the derivative through its update step and the environment. We suspect that this happens because LOLA is a myopic one-step learner, whereas M-FOS considers the discounted returns far in the future. This can be observed in Figure \ref{fig:IPD_plots}. 

Also, note that the NL agent achieves a total score lower than $-2$. This is a lower score than ZD extortion can theoretically make its opponent achieve since blind defection at worst achieves a score of $-2$. Figure \ref{fig:IPD_Viz} shows that M-FOS takes advantage of the fact that the NL agent's gradient updates do not find the optimal response policy in a single step.

\begin{figure}[h]
    \centering
    \includegraphics[width=0.25\textwidth]{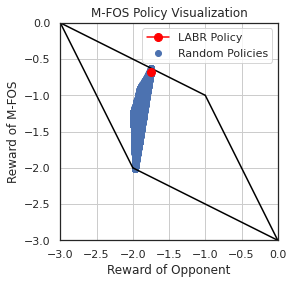}
    \caption{Visualisation of M-FOS v. Look-Ahead Best Response in the IPD. Note that the payoff between the two agents is near-linear and favors the M-FOS agent, indicating ZD extortion.}
    \label{fig:LABR}
\end{figure}

\textit{M-FOS v. Look-Ahead Best Response:} To demonstrate the above point, we train M-FOS against a variant of a naive learner that can observe its opponent's next policy and then plays the best response to it (which is calculated by performing a thousand steps of gradient ascent). Despite the game being symmetric, M-FOS extorts this Look-Ahead Best Response (LABR) agent, achieving an average score of $-0.71$.
Figure \ref{fig:LABR} shows that the policy M-FOS outputs approximates ZD extortion. To the best of our knowledge, M-FOS is the first learning algorithm to discover ZD extortion.

\textbf{M-FOS v. LOLA:} In \citep{foerster_learning_2018}, the authors write that 2nd-order LOLA, which is an agent that takes the derivative through the opponent's LOLA update, does not achieve any incremental gains against an opposing LOLA agent. In other words, a LOLA agent achieves a better score against another LOLA agent than a 2nd-order LOLA agent would, implying that it is difficult to exploit LOLA.

However, M-FOS manages to find a dominating strategy against LOLA ($-0.73$ / $-2.09$). To the best of our knowledge, M-FOS is the first learning algorithm to exploit the LOLA update.

\textbf{M-FOS v. M-MAML:} M-MAML seems to have generally learned to initialize with values close to TFT (see Appendix Section~\ref{sec:mmaml-plot}). This initialisation allows it to achieve favorable results against all algorithms except M-FOS ($-0.67$ / $-1.86$), which learns to exploit it in Figure \ref{fig:IPD_plots}.

\begin{figure}[h]
    \centering
    \includegraphics[width=0.4\textwidth]{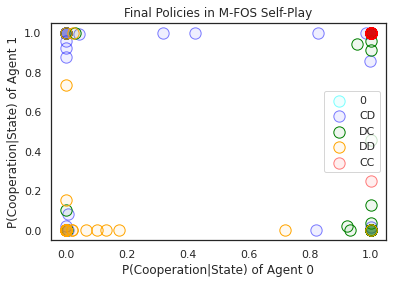}
    \caption{Visualisation of 32 Final Episode Policies in M-FOS v. M-FOS in the IPD
    \vspace{-1.0\baselineskip}
    }
    \label{fig:MFOS_SELF}
\end{figure}

\textbf{M-FOS v. M-FOS:} We arrive at a cooperative score when M-FOS is trained against other M-FOS agents using the meta-self-play training scheme from above. When viewing the final policies played against each other, we observe that M-FOS has largely arrived at TFT, as seen in Figure \ref{fig:MFOS_SELF}. To the best of our knowledge, M-FOS is the first learning algorithm to arrive at TFT in the IPD against itself without using higher-order derivatives, access to the opponent's rewards, or specific hand-coding of TFT-like behaviour.

\subsection{Other Matrix Games}

\begin{figure*}[t!]
    \centering
    \includegraphics[width=0.9\textwidth]{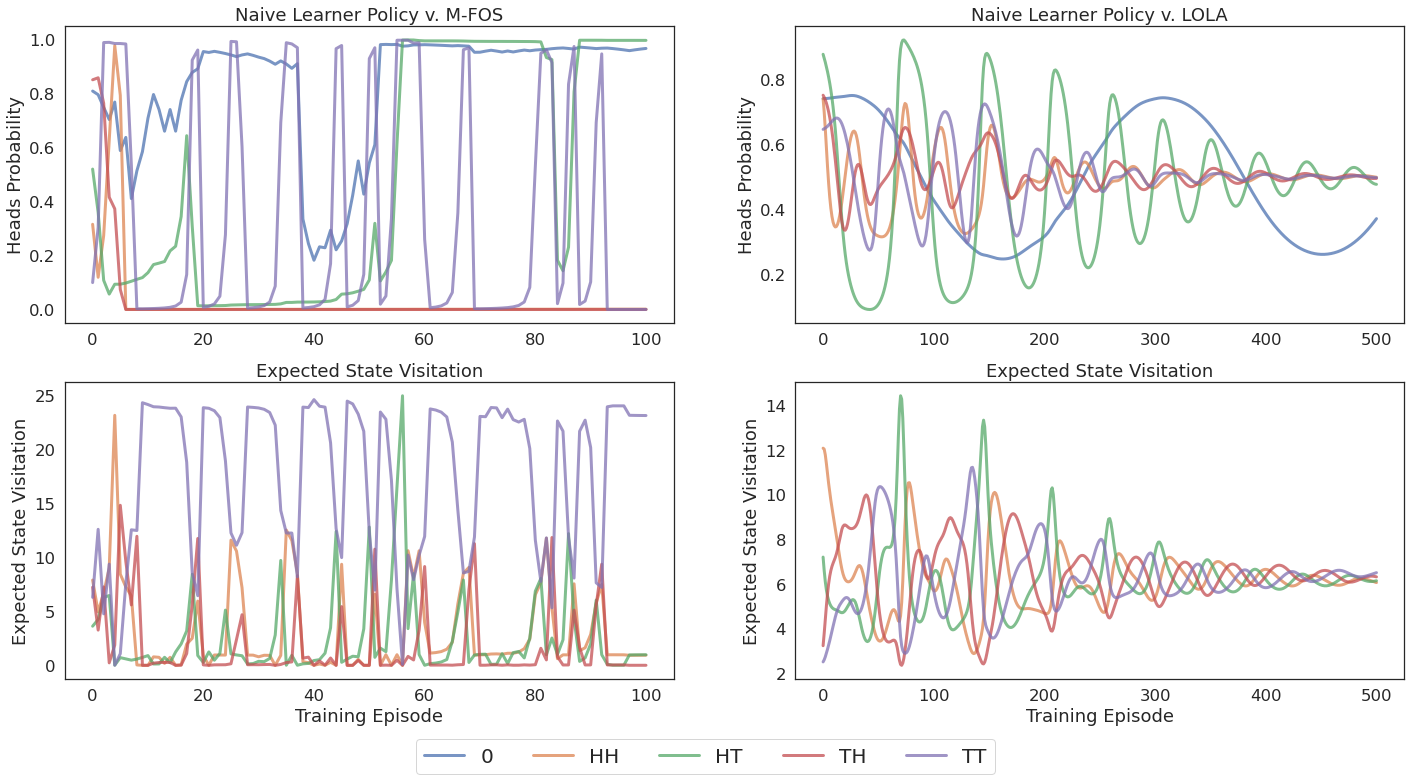}
    \vspace{-10pt}
    \caption{Visualisation of M-FOS's long-term shaping LOLA's and myopic strategy in the Iterated Matching Pennies environment. Note how LOLA converges to the nash equilibrium, resulting in zero reward for both agents, while M-FOS continually drags the naive learner's policy to exploitable states.
    \vspace{-1.0\baselineskip}
    }
    \label{fig:IMP_plots}
\end{figure*}

\begin{table}[hbt!]
\vspace{-1.0\baselineskip}
\caption{Head-to-head results of each learning algorithm in Iterated Matching Pennies.}
\begin{tabular}{l|llll}
      & M-FOS              & NL              & LOLA            & M-MAML \\ \hline
M-FOS  & \textbf{0.0} & \textbf{0.20} & \textbf{0.19} & \textbf{0.22}     \\ \hline
NL     & -0.20          & 0.0           & -0.02           & -0.01     \\ \hline
LOLA   & -0.19          & 0.02           & 0.0           & 0.02     \\ \hline
M-MAML & -0.22             & 0.01              & -0.02              & 0.0     \\ \hline
\end{tabular}
\vspace{-1.0\baselineskip}
\label{table:IMP_results}
\end{table}

\vspace{0.5\baselineskip}
\textbf{IMP:} In IMP, M-FOS once again outperforms other baseline methods. In particular, by examining how M-FOS exploits a naive learner compared to how LOLA does so, we observe that LOLA is myopic compared to M-FOS. In Figure \ref{fig:IMP_plots}, LOLA gradually approaches the nash equilibrium against a naive learner in order to avoid being exploited by its opponent. In contrast, M-FOS cyclically shapes the naive learner's policy to continuously exploit it while staying one step ahead.

\begin{table}[hbt!]
\vspace{-1.0\baselineskip}
\caption{Head-to-head results of each learning algorithm in the Chicken Game. The results of an M-FOS meta-policy that learns an initial policy is in parantheses.}
\label{table:chicken_results}
\begin{tabular}{l|llll}
      & M-FOS              & NL              & LOLA            & M-MAML \\ \hline
M-FOS  & \textbf{-0.01} & \textbf{0.97} & -0.94[\textbf{0.5}] &    \textbf{0.86}   \\ \hline
NL     & -1.03          & -0.0           & -0.97           & -0.27     \\ \hline
LOLA   & \textbf{0.87}[-1.5]       & 0.94           & -85.96           & 0.40     \\ \hline
M-MAML & -1.08             & 0.27              & \textbf{-0.42}              & -0.15     \\ \hline
\end{tabular}
\end{table}

\textbf{Chicken Game:} M-FOS performs well against all baselines in the head-to-head in the Chicken Game but achieves a lower score against LOLA. 
Interestingly, LOLA tends to behave in an extortionary manner in the Chicken Game. After one update against most random policies, it attempts to shape the opponent by heading straight (i.e. defecting) with high probability. While this extortionate behavior works against most learning opponents, it leads to catastrophic results in self-play ($-85.96$). This suggests that LOLA will continue to head straight, whether its opponent swerves or heads straight.

Because M-MAML selects its initial policy, it can shape LOLA from the first time step, preventing LOLA from immediately heading straight after its first update. M-FOS, in contrast, is by default forced to a random initialisation. 
However, if we allow M-FOS also to learn an initial policy, it achieves a much higher score against LOLA ($0.5$), far outperforming M-MAML ($-0.42$). 




\subsection{Coin Game}

Prior work \citep{yu2021mbom} has shown that LOLA-DiCE \citep{foerster2018dice} and Meta-MAPG \citep{kim2021meta-mapg} do not achieve significant results in a \textit{simplified} version of coin game with a fully cooperative reward. Because of this, we do not compare to these baselines. We also observe that M-FOS outperforms PPO in head-to-head training while still achieving good performance in self-play. Meanwhile, PPO agents, when trained together, pick up each other's coins indiscriminately, leading to 0 expected reward.


\begin{table}[hbt!]
\centering
\caption{Head-to-head results of M-FOS and PPO in the Coin Game.}

\begin{tabular}{r|rr}
      & M-FOS          & PPO              \\ \hline
M-FOS & \textbf{20.56} & \textbf{44.26} \\ \hline
PPO    & -24.62          & 4.25           \\ \hline
\end{tabular}
\end{table}

\begin{figure}[h]
    \centering
    \includegraphics[width=0.33\textwidth]{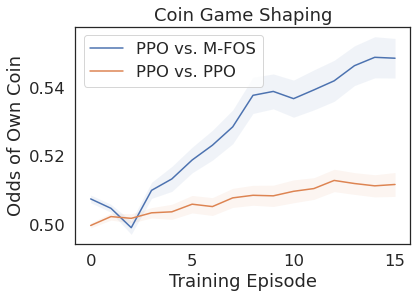}
    \caption{Probability of the PPO agent picking up its own coin across the inner episodes. Note that it is shaped into picking up more of its own coins against the M-FOS agent.
    \vspace{-1.0\baselineskip}
    }
    \label{fig:CoinGameVis}
    
\end{figure}

\section{Conclusion \& Future Work}

In this paper, we presented Model-Free Opponent Shaping (M-FOS) as a simple model-free alternative to popular MAML-like opponent shaping methods, such as LOLA and MMAPG.
Although M-FOS does not use higher-order derivatives and does not have white-box access to its opponent's learning model, it vastly outperforms all tested baselines across several matrix games. 

More specifically, in the IPD, M-FOS achieves several notable results. First, to the best of our knowledge, it is the first learning algorithm to discover ZD extortion, the first learning algorithm that exploits LOLA, and the first learning algorithm to achieve cooperation in self-play without using higher-order derivatives or inconsistent models. Furthermore, it achieves a score higher than mutual cooperation against all tested opponents, while none of the baselines could do so against any single opponent.
We also show that M-FOS can scale to more complex, high-dimensional games and achieve similar results.

In the future, we could generalize M-FOS beyond social dilemmas. For example, M-FOS could shape financial instruments, trading models, recommendation systems, and any system trained on real-world data. 

\section*{Acknowledgements}

Christian Schroeder de Witt is sponsored by the Cooperative AI Foundation. Compute for this project was partially run on Oxford's Advanced Research Cluster (ARC). This work used the Cirrus UK National Tier-2 HPC Service at EPCC funded by the University of Edinburgh and EPSRC (EP/P020267/1). This work was also supported by an Oracle for Research Cloud Grant (19158657).





\raggedbottom
\bibliography{main.bib}
\bibliographystyle{icml2022}

\pagebreak

\onecolumn
\appendix

\section{Detailed Results}\label{sec:detailed-results}

Each experiment is run $10$ times. The inner batch size of each experiment for the matrix games is $4096$.

\begin{table}[hbt!]
\centering
\caption{Head-to-head results of each learning algorithm in IPD, results reported for M-FOS PPO.}
\begin{tabular}{l|llll}
      & M-FOS              & NL              & LOLA            & M-MAML \\ \hline
M-FOS  & -1.01 & -0.51 & -1.03 & -0.84     \\ \hline
NL     & -2.14          & -1.98           & -1.52           & -1.28     \\ \hline
LOLA   & -1.02          & -1.30           & -1.09           & -1.04     \\ \hline
M-MAML & -1.52             & -1.25              & -1.15              & -1.17     \\ \hline
\end{tabular}
\label{table:IPD_PPO}
\end{table}

\begin{table}[hbt!]
\centering
\caption{Head-to-head results of each learning algorithm in IPD, results reported for M-FOS GA.}
\begin{tabular}{l|llll}
      & M-FOS              & NL              & LOLA            & M-MAML \\ \hline
M-FOS  & -- & -0.745 & -0.73 & -0.67     \\ \hline
NL     & -1.69          & -1.98           & -1.52           & -1.28     \\ \hline
LOLA   & -2.09          & -1.30           & -1.09           & -1.04     \\ \hline
M-MAML & -1.86             & -1.25              & -1.15              & -1.17     \\ \hline
\end{tabular}
\label{table:IPD_GA}
\end{table}

\begin{table}[hbt!]
\centering
\caption{Head-to-head results of each learning algorithm in IMP, results reported for M-FOS PPO.}
\begin{tabular}{l|llll}
      & M-FOS              & NL              & LOLA            & M-MAML \\ \hline
M-FOS  & 0.0 & 0.20 & 0.19 & 0.22     \\ \hline
NL     & -0.20          & 0.0           & -0.02           & -0.01     \\ \hline
LOLA   & -0.19          & 0.02           & 0.0           & 0.02     \\ \hline
M-MAML & -0.22             & 0.01              & -0.02              & 0.0     \\ \hline
\end{tabular}
\label{table:IMP_PPO}
\end{table}

\begin{table}[hbt!]
\centering
\caption{Head-to-head results of each learning algorithm in IMP, results reported for M-FOS GA.}
\begin{tabular}{l|llll}
      & M-FOS              & NL              & LOLA            & M-MAML \\ \hline
M-FOS  & -- & 0.13 & 0.10 & 0.17     \\ \hline
NL     & -0.13          & 0.0           & -0.02           & -0.01     \\ \hline
LOLA   & -0.10          & 0.02           & 0.0           & 0.02     \\ \hline
M-MAML & -0.17             & 0.01              & -0.02              & 0.0     \\ \hline
\end{tabular}
\label{table:IMP_GA}
\end{table}

\begin{table}[hbt!]
\centering
\caption{Head-to-head results of each learning algorithm in the Chicken Game. The results of an M-FOS meta-policy that learns an initial policy is in parantheses. Results reported for M-FOS PPO.}
\label{table:chicken_PPO}
\begin{tabular}{l|llll}
      & M-FOS              & NL              & LOLA            & M-MAML \\ \hline
M-FOS  & -0.01 & 0.97 & -0.94[0.5] &    0.85   \\ \hline
NL     & -1.03          & -0.0           & -0.97           & -0.27     \\ \hline
LOLA   & 0.87[-1.5]       & 0.94           & -85.96           & 0.40     \\ \hline
M-MAML & -1.11             & 0.27              & -0.42              & -0.15     \\ \hline
\end{tabular}
\end{table}

\begin{table}[hbt!]
\centering
\caption{Head-to-head results of each learning algorithm in the Chicken Game. The results of an M-FOS meta-policy that learns an initial policy is in parantheses. Results reported for M-FOS GA.}
\label{table:chicken_GA}
\begin{tabular}{l|llll}
      & M-FOS              & NL              & LOLA            & M-MAML \\ \hline
M-FOS  & -- & 0.97 & -0.94[0.5] &    0.86   \\ \hline
NL     & -1.03          & -0.0           & -0.97           & -0.27     \\ \hline
LOLA   & 0.91[-1.5]       & 0.94           & -85.96           & 0.40     \\ \hline
M-MAML & -1.08             & 0.27              & -0.42              & -0.15     \\ \hline
\end{tabular}
\end{table}

\section{M-MAML Initialisations Plot}\label{sec:mmaml-plot}

\begin{figure}[hbt!]
    \centering
    \includegraphics[width=0.5\textwidth]{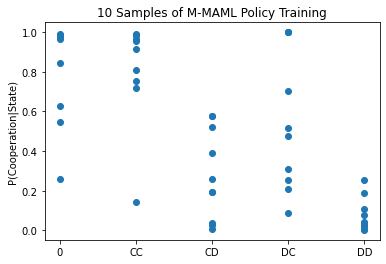}
    \caption{The distribution of probabilities in each state after training $10$ different instances of M-MAML.
    }
    \label{fig:MMAML}
    
\end{figure}

\section{Hyperparameter Details}\label{sec:hyperparameter-details}
We report our hyperparameter values that we used for each of the methods in our experiments:
\subsection{M-FOS}
\begin{table}[H]
\centering
\begin{tabular}{l|l}
Hyperparameter & Value \\ \hline
Number of Actor Hidden Layers & 1 \\
Size of Actor Hidden Layers & [256] \\
Number of Critic Hidden Layers & 1 \\
Size of Critic Hidden Layers & [256] \\
Length of Meta-Episode $T$ & 100 \\
Batch Size $B$ & 4096 \\
Adam Step Size & 0.0002 \\
Number of Epochs & 4 \\
Outer Discount Factor $\gamma$ & 0.99 \\
PPO Clipping $\epsilon$ & 0.2 \\
Entropy Coefficient & 0.01 \\
\end{tabular}
\caption{PPO for IPD, IMP, and Chicken Game}
\end{table}

\begin{table}[H]
\centering
\begin{tabular}{l|l}
Hyperparameter & Value \\ \hline
Number of Hidden Layers & 1 \\
Size of Hidden Layers & [256] \\
Number of Species $N$ & 2048 \\
Batch Size $B$ & 128 \\
Length of Meta-Episode $T$ & 100 \\
Noise Std Dev $\sigma$ & 2.0 \\
Number of Elites $E$ & 1 \\
\end{tabular}
\caption{Genetic Algorithm for IPD, IMP, and Chicken Game}
\end{table}

\begin{table}[H]
\centering
\begin{tabular}{l|l}
Hyperparameter & Value \\ \hline
Number of Conv Layers & 2 \\
Output Channels of Conv Layers & [16, 16] \\
Kernel Sizes of Conv Layers & [[3, 3], [3, 3]] \\
Strides of Conv Layers & [1, 1] \\
Number of Linear Layers & 1 \\ 
Size of Linear Layer & [16] \\
Number of GRUs & 1 \\
Size of GRUs & [16] \\
Length of Meta-Episode $T$ & 16 \\
Length of Inner Episode & 16 \\
Batch Size $B$ & 512 \\
Adam Step Size & 0.0002 \\
Number of Epochs & 16 \\
Outer Discount Factor $\gamma$ & 0.99 \\
PPO Clipping $\epsilon$ & 0.2 \\
Entropy Coefficient & 0.01 \\
\end{tabular}
\caption{PPO For Coin Game. The Actor, Critic, and Meta-Policy have the same network architecture but do not share weights.}
\end{table}

\subsection{Environments}

\begin{table}[H]
\centering
\begin{tabular}{l|l}
Hyperparameter & Value \\ \hline
Inner Gamma $\gamma$ & 0.96 \\
Learning Rate $\alpha$ & 1 \\
M-MAML Adam Learning Rate & 0.05 \\
\end{tabular}
\caption{Hyperparameters for IPD Environment}
\end{table}

\begin{table}[H]
\centering
\begin{tabular}{l|l}
Hyperparameter & Value \\ \hline
Inner Gamma $\gamma$ & 0.96 \\
Learning Rate $\alpha$ & 0.1 \\
M-MAML Adam Learning Rate & 0.05 \\
\end{tabular}
\caption{Hyperparameters for IMP Environment}
\end{table}

\begin{table}[H]
\centering
\begin{tabular}{l|l}
Hyperparameter & Value \\ \hline
Learning Rate $\alpha$ & 1 \\
M-MAML Adam Learning Rate & 0.05 \\
\end{tabular}
\caption{Hyperparameters for Chicken Environment}
\end{table}

\begin{table}[H]
\centering
\begin{tabular}{l|l}
Hyperparameter & Value \\ \hline
Number of Conv Layers & 2 \\
Output Channels of Conv Layers & [16, 16] \\
Kernel Sizes of Conv Layers & [[3, 3], [3, 3]] \\
Strides of Conv Layers & [1, 1] \\
Number of Linear Layers & 1 \\ 
Size of Linear Layer & [16] \\
Adam Step Size & 0.005 \\
Number of Epochs & 80 \\ 
PPO Clipping $\epsilon$ & 0.2 \\
Entropy Coefficient & 0.01 \\
Discount Factor $\gamma$ & 0.96 \\
Length of Inner Episode & 16 \\
\end{tabular}
\caption{Hyperparameters for Coin Game Environment and Naive Learner. The Actor and Critic share the same architecture but do not share weights.}
\end{table}

\end{document}